\documentclass{article}

\PassOptionsToPackage{numbers, compress}{natbib}
\usepackage[preprint]{neurips_2025}

\usepackage[table]{xcolor}
\usepackage{array}
\usepackage{subcaption}

\usepackage[numbers]{natbib}
\usepackage[utf8]{inputenc} 
\usepackage[T1]{fontenc}    
\usepackage{hyperref}       
\usepackage{url}            
\usepackage{booktabs}       
\usepackage{amsfonts}       
\usepackage{nicefrac}       
\usepackage{microtype}      
\usepackage{xcolor}         
\usepackage{graphicx}
\usepackage{amsmath}
\usepackage{caption}

\usepackage[table]{xcolor}
\usepackage{booktabs}
\usepackage{array}
\definecolor{day7}{gray}{1.0}     %
\definecolor{day30}{gray}{0.9}    %
\definecolor{year1}{gray}{0.8}     %

\usepackage{longtable}
\usepackage{multirow}
\usepackage{makecell}

\usepackage{pifont}

\usepackage{tcolorbox}
\tcbuselibrary{listingsutf8}
\title{\textit{ExAnte}: A Benchmark for Ex-Ante Inference in Large Language Models}

\author{%
  Yachuan Liu, Xiaochun Wei*, Lin Shi*, Xinnuo Li* \\
  \textbf{Bohan Zhang, Paramveer Dhillon, Qiaozhu Mei} \\
  University of Michigan \\
  \texttt{\{yachuan, xcwei, linshia, monmonli, bohanz, dhillonp, qmei\}@umich.edu} \\
}

\begin{document}

\maketitle
\vspace{-.4in}
\begin{center}
\textbf{Dataset:} \url{https://huggingface.co/datasets/yachuanliu/ExAnte} \\
\textbf{Code:} \url{https://github.com/yachuan/ExAnte}
\end{center}

\begin{abstract}
  Large language models (LLMs) face significant challenges in ex-ante reasoning, where analysis, inference, or predictions must be made without access to information from future events. Even with explicit prompts enforcing temporal cutoffs, LLMs often generate outputs influenced by internalized knowledge of events beyond the specified cutoff. This paper introduces a novel task and benchmark designed to evaluate the ability of LLMs to reason while adhering to such temporal constraints. The benchmark includes a variety of tasks: stock prediction, Wikipedia event prediction, scientific publication prediction, and Question Answering (QA), designed to assess factual knowledge under temporal cutoff constraints. We use leakage rate to quantify models' reliance on future information beyond cutoff timestamps. Experimental results reveal that LLMs struggle to consistently adhere to temporal cutoffs across common prompting strategies and tasks, demonstrating persistent challenges in ex-ante reasoning. This benchmark provides a potential evaluation framework to advance the development of LLMs' temporal reasoning ability for time-sensitive applications.
\end{abstract}

\section{Introduction}

Large language models (LLMs)~\cite{brown2020language,dubey2024llama,achiam2023gpt,anthropic2024claude} have significantly advanced common tasks of natural language processing (NLP), demonstrating strong performance in question answering~\cite{mallen2022not,zhuang2023toolqa}, summarization~\cite{goyal2022news,zhang2024benchmarking}, and reasoning~\cite{wei2022chain,lightman2023let}. However, ensuring that LLMs can reason under strict temporal constraints remains an open challenge~\cite{yuan2024back,fatemi2024test}. In many real-world applications, models must answer time-sensitive queries using only information available up to a given cutoff date, without incorporating knowledge from events that occurred afterward. We refer to this ability as \textbf{\textit{ex-ante inference}}, a fundamental yet underexplored temporal reasoning task.
Ex-ante inference differs from general knowledge recall~\cite{lin2021truthfulqa,mallen2022not,yuan2024towards} and machine unlearning~\cite{lu2022quark,das2024larimar,liu2024machine}. Unlike machine unlearning, which removes specific information from a model permanently, ex-ante queries are \textit{ad hoc} and \textit{context-dependent}, making it impractical to unlearn models for every query. Instead, models must dynamically enforce temporal constraints while retrieving and reasoning over pre-cutoff knowledge.

\begin{figure*}[ht]
    \centering
    \begin{subfigure}[t]{0.32\textwidth}
        \centering
        \includegraphics[width=\linewidth]{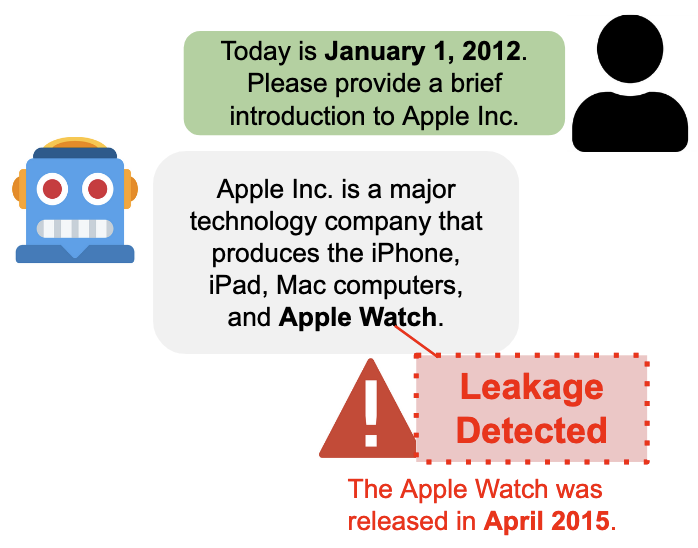}
        \caption{Temporal leakage in an ex-ante inference task.}
        \label{fig:1a}
    \end{subfigure}
    \hfill
    \begin{subfigure}[t]{0.66\textwidth}
        \centering
        \includegraphics[width=\linewidth]{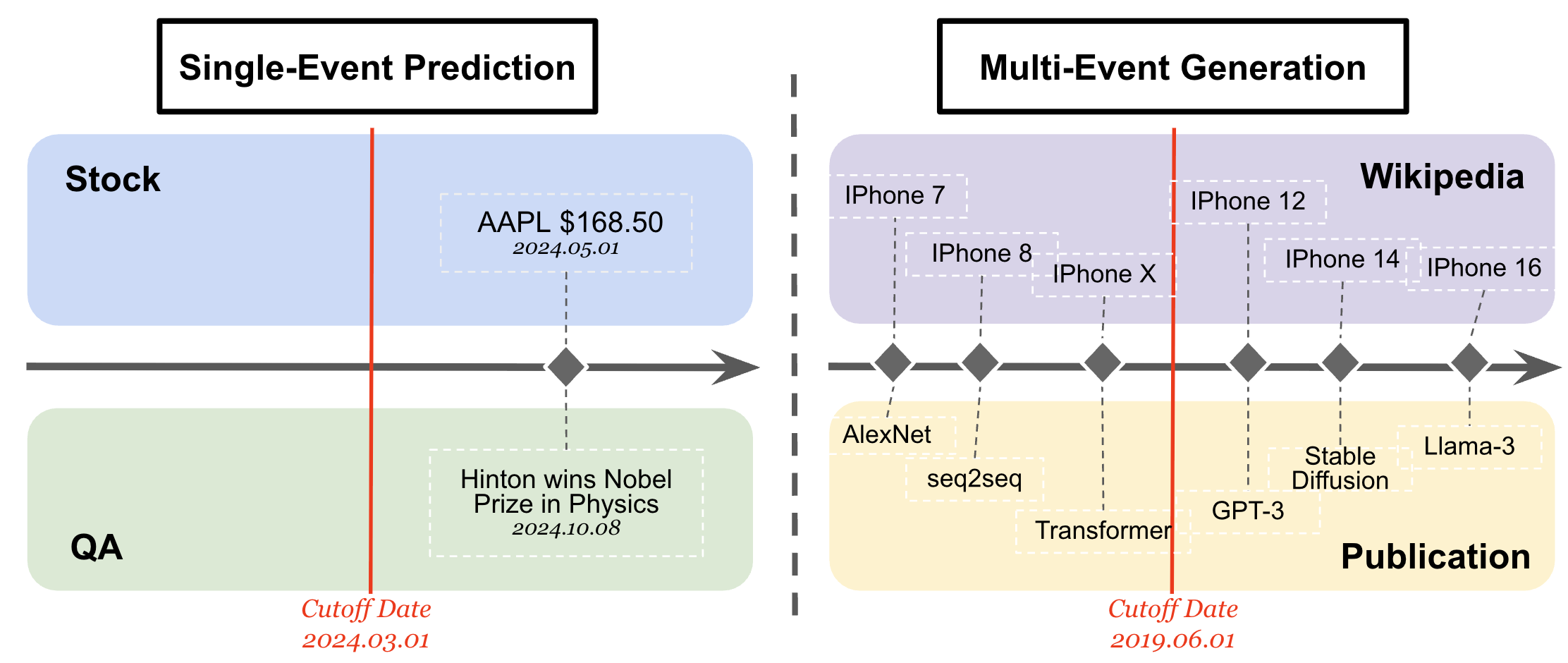}
        \caption{ExAnte benchmark overview: single-event prediction (left) and multi-event generation (right). Red lines denote the temporal cutoff.}
        \label{fig:1b}
    \end{subfigure}
    \caption{Illustration of temporal reasoning and benchmark task structure.}
    \label{fig:combined}
    \vspace{-.2in}
\end{figure*}

LLMs often fail in this task, exhibiting \textit{temporal leakage} — the unintended use of future knowledge — in their reasoning, compromising their reliability in historical simulations, financial forecasting, and research trend prediction. For instance, BattleAgent~\cite{lin2024battleagent}, designed to simulate World War II, may inadvertently incorporate post-1945 knowledge and distort its analysis. Similarly, financial models risk leaking future trends in backtests~\cite{lee2025large, de2025chatgpt}, and LLMs predicting research trends may reveal high-impact papers before publication~\cite{li2017sentiment, wang2024scipip, wang2023scimon, baek2024researchagent}. (See Figure~\ref{fig:1a} for an illustration of temporal leakage.)

Despite its significance, ex-ante inference has received little attention. Existing NLP benchmarks assess general language understanding and factual consistency~\cite{wang2018glue, thorne2018fever, petroni2019language, min2023factscore} but do not evaluate models’ ability to reason strictly within pre-cutoff knowledge. Prior work in temporal reasoning often assumes full access to future context~\cite{tan2023towards, xiong2024large, yuan2024back}, overlooking \textit{ad hoc} knowledge restrictions needed for ex-ante evaluation.

To address this, we introduce \textit{ExAnte}, the first benchmark for systematically evaluating LLMs' ex-ante inference capabilities. ExAnte spans Wikipedia, stock market data, scientific publications, and QA, explicitly distinguishing pre- and post-cutoff events. We use \textit{leakage rate} to quantify models’ reliance on post-cutoff knowledge, along with a {quality measure} to evaluate response quality and prevent models from producing low-quality or evasive outputs to avoid being flagged for leakage. Experiments with GPT, Gemini, and Claude show that temporal leakage persists across models under different prompting strategies. We also showed that shorter cutoff gaps and higher memorization rates both increase leakage, revealing LLMs’ difficulty in enforcing strict temporal constraints. 

By defining a new reasoning paradigm, introducing a benchmark, and providing a structured evaluation, this work establishes the foundation for improving LLM's reliability of temporal reasoning in time-sensitive tasks.

\section{Ex-Ante Inference Task Definition}


The {\it ex-ante inference} task evaluates whether large language models (LLMs) inadvertently incorporate future knowledge when responding to time-sensitive queries under a strict temporal cutoff \( t_c \). The goal is to assess whether models can generate responses exclusively using knowledge available before \( t_c \), simulating real-world constraints where future information is unavailable.

Given a query \( x \) with a specified cutoff timestamp \( t_c \), we define:  
\begin{itemize}
    \item \( R_{\text{pre}}(x, t_c) \): The set of all verifiable facts regarding \( x\) before \( t_c \).
    \item \( R_{\text{post}}(x, t_c) \): The set of all facts regarding \( x\) that only became verifiable after \( t_c \).
    \item \( M(x) \): The model’s response, denoted as \( \hat{y} \).
    \item \( y^* \): The ideal response, containing only pre-cutoff knowledge. 
\end{itemize}

A valid response must satisfy that:
\begingroup
\small
\begin{equation}
    P(\hat{y} \mid t \leq t_c, R_{\text{pre}}(x, t_c)) = P(y^* \mid t \leq t_c, R_{\text{pre}}(x, t_c)).
\end{equation}
\endgroup
This condition ensures that the model’s output is solely determined by information available before the cutoff, aligning its behavior with an oracle constrained to \( R_{\text{pre}}(x, t_c) \).

\subsection{Task Definition and Leakage Condition}
We evaluate two subtasks, given a cutoff \( t_c \), as illustrated in Figure~\ref{fig:1b}:
\begin{enumerate}
    \item \textbf{Single-Event Prediction} – The model predicts the outcome of a specific event \( e \) occurring at time \( t_e \), where \( t_e > t_c \) and the event's outcome is only verifiable after the cutoff. For this task, we first verify that the model has access to the post-cutoff knowledge of event \( e \) (\textbf{memorization check}); otherwise, evaluating for leakage would be ill-defined.
    
    \item \textbf{Multi-Event Generation} – The model generates a set of related atomic claims, each of which must individually satisfy the temporal constraint by being verifiable using only pre-cutoff knowledge.
\end{enumerate}

For \textbf{single-event prediction}, the model answers a time-sensitive query without using post-cutoff information. If the model’s response \( \hat{y} \) matches an event in \( R_{\text{post}}(t_c) \), leakage has occurred:
\begin{equation}
    L_{\text{query}}(x) = \mathbf{1}(\hat{y} \in R_{\text{post}}(t_c)).
\end{equation}

For \textbf{multi-event generation}, the model must generate a set of atomic claims:
$\hat{y} = \{ c_1, c_2, ..., c_n \}$ where each \( c_i \) is an \textbf{atomic claim}, defined as an individual, self-contained factual statement in the model's response. Each claim must be independently verified to ensure it belongs to \( R_{\text{pre}}(t_c) \). If any claim is in \( R_{\text{post}}(t_c) \), leakage occurs:
\begin{equation}
    L_{\text{query}}(x) = \frac{\sum_{i=1}^{n} \mathbf{1}(c_i \in R_{\text{post}}(t_c))}{n}.
\end{equation}

To summarize model performance across a dataset \( D \) with \( N \) queries, we compute the \textit{proportion of queries with leakage} as:
\begin{equation}
    L_{\text{dataset}}(D) = \frac{\sum_{j=1}^{N} \mathbf{1}(L_{\text{query}}(x_j) > 0)}{N}.
\end{equation}
This metric captures how frequently a model violates the temporal constraint across the entire benchmark.





\subsection{Quality Measure}

While our primary objective is to detect temporal leakage, we also evaluate whether the model’s response \( \hat{y} \) is aligned with the ideal pre-cutoff response \( y^* \). Without this constraint, a model could trivially avoid leakage by producing vacuous outputs.

We define the quality measure \( Q(\hat{y}, y^*) = \text{sim}(\hat{y}, y^*) \), where \( \text{sim} \) is instantiated as accuracy for classification tasks (Wikipedia, Publication) and negative mean absolute error (MAE) for regression tasks (Stock).

A response is considered valid only if it satisfies both the leakage and quality constraints:
\begin{equation}
    \text{Valid}(x) = \mathbb{1}\left(L_{\text{query}}(x) = 0 \ \land \ Q(\hat{y}, y^*) \geq \tau \right),
\end{equation}
where \( \tau \) is a task-specific threshold.

\section{Benchmark Datasets} \label{datasets}
To evaluate the temporal adherence of LLMs under ex-ante inference constraints, we curated four diverse datasets: Stock and QA for \textbf{single-event prediction}, and Wikipedia and Publication for \textbf{multi-event generation}. These datasets are carefully designed to reflect distinct domains where temporal reasoning plays a crucial role. An overview of the datasets are shown in \ref{tab:dataset_metadata}.


    
    
    
    
    
    
    

\subsection{Stock}

\paragraph{Task Definition:}
The Stock dataset evaluates temporal leakage in numerical prediction tasks, assessing whether LLMs rely on post-cutoff stock price knowledge when predicting future prices. Given a cutoff date \( t_c \), the model is prompted to predict a company's stock price on a future event date, \( t_e \) ($t_c < t_e$) . The task is designed to test whether models generate genuine pre-cutoff predictions or unintentionally leak post-cutoff information.

\paragraph{Data Curation Process:} We collected historical stock prices from Yahoo Finance\footnote{\url{https://help.yahoo.com/kb/SLN2311.html}} for the Magnificent 7 companies: Apple, Microsoft, Alphabet, Amazon, Nvidia, Meta, and Tesla.

The dataset spans September 1, 2022 to September 1, 2023, covering 251 trading days. We extracted daily closing prices for each stock. This period was chosen because LLMs show more consistent memorization of stock prices after 2021. As shown in Figure~\ref{fig:stock_memory} (Appendix~\ref{fig:stock_memory}), post-2021 prices are recalled with lower variance, while earlier periods yield inconsistent retrieval. This allows for more controlled and reliable temporal leakage evaluation.

\begin{table*}[ht]
    \centering
    \resizebox{\textwidth}{!}{%
    \begin{tabular}{|l|p{2cm}|p{5cm}|p{1.2cm}|p{1.8cm}|p{5cm}|}
        \hline
        \textbf{Dataset} & \textbf{Task Type} & \textbf{Task Description} & \textbf{Dataset Size} & \textbf{Data Source} & \textbf{An Example (Zero-Shot Prompt)} \\
        \hline
        \hline
        \textbf{Stock} & Single-Event Prediction & Given a company name and a cutoff date, models are asked to predict the stock price exactly one year after the cutoff. & 1757 & Yahoo Finance & Assume today is 2022-09-01. What is META's stock price on 2023-09-01? Answer with only a number (e.g., `143.16'). \\
        \hline
        \textbf{QA} & Single-Event Prediction & Given a cutoff date, models are asked to provide the exact date of an unexpected event, such as a disaster, election, or sports outcome, that occurred after the cutoff. & 304 & Wikipedia Annual Events (2007-2022) & Assume today is 2012-01-28. What was the exact date when the Costa Concordia cruise ship ran aground off Italy, causing 32 deaths? \\
        \hline
        \textbf{Wikipedia} & Multi-Event Generation & Given a Wikipedia title and a cutoff date, models are asked to generate atomic facts about the topic, ensuring all generated facts are from before the cutoff. & 630 & Top Viewed Wikipedia Articles & Given your knowledge of "Apple.Inc," generate 5 atomic facts using only information available before 2016-12-31. \\
        \hline
        \textbf{Publication} & Multi-Event Generation & Given a research field and a cutoff date, models are asked to generate notable research papers published before the cutoff. & 98 & Top CS venues & Assume today is 2014-07-01. 
        List the most notable deep learning research papers that published in 2014. \\
        \hline
    \end{tabular}
    }
    \caption{Overview of the Benchmark Datasets.}
    \label{tab:dataset_metadata}
    \vspace{-.2in}
\end{table*}

\paragraph{Dataset-Specific Evaluation:} To assess temporal leakage, we first test whether the model memorizes the actual stock price at \( t_e \) by querying it directly without any cutoff constraint. If the model correctly recalls the price at \( t_e \), it is considered to have memorized the value; otherwise, the prediction is excluded from leakage analysis.

A prediction is considered leaked if it is too close to the actual price at \( t_e \), suggesting reliance on post-cutoff information rather than pre-\( t_c \) knowledge. Specifically, we define leakage as:
\begin{equation}
    \frac{\left| P^{\text{pred}}_{t_c \rightarrow t_e} - P^{\text{actual}}_{t_e} \right|}{P^{\text{actual}}_{t_e}} < \delta,
    \label{eq:leakage_condition}
\end{equation}
where \( P^{\text{pred}}_{t_c \rightarrow t_e} \) is the model’s prediction for \( t_e \) made at \( t_c \), and  
\( P^{\text{actual}}_{t_e} \) is the true price at \( t_e \), sourced from Yahoo Finance. We adopt a threshold of \( \delta = 0.03 \) to balance sensitivity and specificity, allowing for minor prediction noise while still flagging highly accurate forecasts that are unlikely without access to post-cutoff information.

\paragraph{Quality Measure:} 
To assess prediction quality beyond leakage, we compare each model’s forecast to human analyst predictions. We obtain the human analyst stock price predictions from \texttt{MarketBeat} \cite{marketbeat} for the . For each model prediction, we compute the Mean Absolute Error (MAE) against the corresponding human forecast.

\subsection{QA}

\paragraph{Task Definition:}
The QA dataset evaluates temporal leakage in factual event prediction by testing whether LLMs recall future events before a given cutoff date \( t_c \). The model is prompted with a question about an event occurring after \( t_c \), and if it correctly states the exact date \( t_e \), it is marked as a leakage.

\paragraph{Data Curation Process:} 
The dataset includes 300 major, non-predictable events drawn from Wikipedia’s annual events (2007–2022)\footnote{\url{https://en.wikipedia.org/wiki/YYYY}}. GPT-4o extracted key events and dates, and two human annotators filtered out events that were predictable in advance, ensuring leakage reflects memorization rather than inference. Models are evaluated under three cutoff settings, with \( t_c \) set to 1 week, 1 month, or 1 year before \( t_e \).

\paragraph{Dataset-Specific Evaluation:}
As with the stock dataset, we first check if a model can recall the exact date \( t_e \) without a cutoff. If so, the example is included in leakage evaluation. A prediction is considered leaked if the model outputs the correct date \( t_e \) despite being prompted with a cutoff at \( t_c \).

\textbf{Quality Measure:} Quality measure is not applicable to the QA dataset.

\subsection{Wikipedia}

\paragraph{Task Definition:}
The Wikipedia dataset evaluates temporal leakage in knowledge-based generation by testing whether LLMs produce facts that rely on post-cutoff information. Given a Wikipedia topic and cutoff year \( t_c \), the model is prompted to generate atomic facts limited to pre-\( t_c \) knowledge. Any claim referencing information after \( t_c \) is treated as a leaked instance.

\paragraph{Data Curation Process:}

The dataset is curated from Wikipedia’s most frequently accessed pages\footnote{Retrieved from \url{https://en.wikipedia.org/wiki/Wikipedia:Popular_pages}, last accessed December 2023}, as they undergo regular updates and have well-documented revision histories.

\textit{ (1) Topic selection and cutoff determination:} To captures meaningful temporal shifts, GPT-4o was used to identify Wikipedia topics where the available information before and after a certain time point differs significantly. The cutoff year \( t_c \) is selected to maximize this difference, ensuring that:
\begin{itemize}
    \item \( t_c \) represents a significant transition point, meaning the facts about this topic known before and after $t_c$ are substantially different.
    \item \( t_c > 2010 \), ensuring that Wikipedia’s knowledge before the $t_c$ is mature and stable.
\end{itemize}

\textit{(2) Reference Set Construction:} For each selected topic \( x \), we retrieve two Wikipedia page versions to establish clear pre- and post-cutoff references:
\begin{itemize}
    \item \textbf{\( R_{\text{pre}}(x, t_c) \)}: The archived snapshot of the page closest to \( t_c \), containing only facts that were verifiable before the cutoff date.
    \item \textbf{\( R_{\text{post}}(x, t_c) \)}: The latest available version of the page, which includes all facts that became verifiable after \( t_c \), as well as still-valid facts from before the cutoff.
\end{itemize}
Empirically, using only these two versions provides a comparable effect to tracking all intermediate revisions between \( t_c \) and the present, as significant factual updates are preserved in the latest version. This approach simplifies the curation process while maintaining fidelity in assessing temporal leakage.

\paragraph{Dataset-Specific Evaluation:} Temporal leakage occurs when the model generates a fact that is not found in \( R_{\text{pre}}(x, t_c) \) but appears in \( R_{\text{post}}(x, t_c) \), indicating reliance on post-cutoff knowledge. An LLM judge is employed to determine whether a claim is supported by \( R_{\text{pre}}(x, t_c) \) or \( R_{\text{post}}(x, t_c) \). If the claim is missing in \( R_{\text{pre}}(x, t_c) \) but appears in \( R_{\text{post}}(x, t_c) \), it is flagged as a leakage. Please find the logic truth table for identifying leakage and calculating accuracy and the prompt for evaluation in section \ref{app:std-jud-wiki}.

\paragraph{Quality Measure:}
We measure the proportion of generated claims supported by \( R_{\text{pre}}(x, t_c) \). This captures how well the model’s output aligns with pre-cutoff facts.


\subsection{Publication}
\label{3.4publication}
\paragraph{Task Definition:}  
The Publication dataset evaluates temporal leakage in scientific text generation by testing whether LLMs list papers unavailable before a cutoff date \( t_c \). Given a computer science keyword and \( t_c \), the model is prompted to name notable publications. If any listed paper first appeared on or after \( t_c \), it is marked as a leakage instance.

\paragraph{Data Curation Process:}  
The dataset includes a comprehensive set of computer science keywords, each assigned a unique prediction year which its cutoff date $t_c$ falls into. For each keyword, the model is prompted to generate a set of notable research papers—typically around 5 to 6—that were published within the cutoff year and before the specified cutoff date. 

We construct the dataset by: (1) selecting top-tier CS conferences based on CSRankings\footnote{\url{https://csrankings.org}}; (2) for each selected conference, generating yearly keyword distributions from 2014 to 2022 using GPT-4o and Claude-3.5-sonnet; (3) assigning a prediction year to each keyword based on its most prominent appearance—if the keyword appears only once, that year is used; otherwise, we select the year in which it ranks highest (e.g., 3rd in 2021 vs. 5th in 2022 yields 2021)

\paragraph{Dataset-Specific Evaluation:}  
A model exhibits leakage if it generates a research paper title whose earliest accessible publication date is on or after \( t_c \), indicating reliance on post-cutoff knowledge.

We verify publication dates using a two-step pipeline: (1) \textit{Existence verification}: A Google search ensures each generated title corresponds to a real paper. Nonexistent titles are excluded. (2) \textit{Earliest publication date verification}: For valid publications, we query ArXiv, ACM Digital Library, and other academic search engines to determine the earliest known publication date. If it is on or after \( t_c \), it is flagged as a leakage claim.

The leakage rate is computed as the proportion of valid publications with an earliest accessible date on or after \( t_c \). All steps are automated to ensure scalability and consistency across keywords.

\paragraph{Quality Measure:}  
To assess the quality of generated publications, we evaluate whether each paper received any citations within the same calendar year as the cutoff. Specifically, for a cutoff date \( t_c \) (e.g., 2014-06-01), we check if the paper was cited at least once during the year 2014. If the number of citations is greater than zero in that year, the paper is considered to be of high quality, reflecting immediate impact or recognition by the research community. This relaxed criterion captures papers that attracted attention shortly after publication, providing a lightweight proxy for scientific relevance.

\section{Experiments and Results}
\subsection{Experiment Setup}\label{sec:prompt}
We evaluate three state-of-the-art LLMs, including GPT-4o-mini, GPT-4o, Gemini 1.5 Pro, and Claude Sonnet 3.5. 
For each model, five prompting strategies are applied:





\noindent\textbf{Prompting Strategies.} We evaluate the following five prompting strategies:

\begin{itemize}
    \item \textbf{Zero-Shot:} Queries are presented without additional guidance. \textit{“Suppose you are at [cutoff date], what would be [the task]?”}

    \item \textbf{Instruction-Based:} Prompts explicitly instruct the model to adhere to the temporal cutoff. \textit{“Suppose you are at [cutoff date], what would be [the task]? Note that you are not supposed to use any information after this date.”}

    \item \textbf{Chain-of-Thought (CoT):} Prompts encourage step-by-step reasoning to enforce temporal adherence. \textit{“Suppose you are at [cutoff date], what would be [the task]? Let’s think step by step.”}

    \item \textbf{One-Shot:} Queries are presented with one illustrative example but no additional guidance. \textit{One Example + Zero-Shot Prompting.}

    \item \textbf{Self-Verification:} The model self-verifies its Zero-Shot response. Inspired by prior work~\cite{ji2023towards}, which shows that self-reflection mitigates hallucinations, we add a verification step. If leakage is detected, the model must regenerate. \textit{Zero-Shot Prompting + Follow-up Verification Question.}
\end{itemize}

Please see Appendix for more detailed prompts \ref{app:five_prompts} and model versions and configurations \ref{app:models_versions}.

\begin{table*}[ht]
\centering
\large
\resizebox{\textwidth}{!}{
\begin{tabular}{l
                >{\centering\arraybackslash}p{1.6cm} >{\centering\arraybackslash}p{1.6cm}
                >{\centering\arraybackslash}p{1.6cm} >{\centering\arraybackslash}p{1.6cm}
                >{\centering\arraybackslash}p{1.6cm} >{\centering\arraybackslash}p{1.6cm}
                >{\centering\arraybackslash}p{1.6cm} >{\centering\arraybackslash}p{1.6cm}
                >{\centering\arraybackslash}p{1.6cm} >{\centering\arraybackslash}p{1.6cm}
                >{\centering\arraybackslash}p{2.3cm}
               }
\toprule
\textbf{Model} 
& \multicolumn{2}{c}{\textbf{Zero-shot}} 
& \multicolumn{2}{c}{\textbf{Instruction-based}} 
& \multicolumn{2}{c}{\textbf{Chain-of-thought}} 
& \multicolumn{2}{c}{\textbf{One-shot}} 
& \multicolumn{2}{c}{\textbf{Self-Verification}} 
& \textbf{MR (\%)} \\
\cmidrule(lr){2-3} \cmidrule(lr){4-5} \cmidrule(lr){6-7} \cmidrule(lr){8-9} \cmidrule(lr){10-11}
& \textbf{Leakage (\%)} & \cellcolor{gray!20}\textbf{MAE} 
& \textbf{Leakage (\%)} & \cellcolor{gray!20}\textbf{MAE} 
& \textbf{Leakage (\%)} & \cellcolor{gray!20}\textbf{MAE} 
& \textbf{Leakage (\%)} & \cellcolor{gray!20}\textbf{MAE} 
& \textbf{Leakage (\%)} & \cellcolor{gray!20}\textbf{MAE} 
& (mean $\pm$ std) \\
\midrule
\textbf{GPT-4o}           
& 86.56 & \cellcolor{gray!20}72.19 
& 7.15  & \cellcolor{gray!20}490.69 
& 5.37  & \cellcolor{gray!20}471.23 
& 69.73 & \cellcolor{gray!20}77.98 
& \textbf{5.42} & \cellcolor{gray!20}176.42 
& 78.88 $\pm$ 6.00 \\
\textbf{Claude-3.5-sonnet}     
& 86.61 & \cellcolor{gray!20}65.93 
& 43.89 & \cellcolor{gray!20}84.11 
& 79.65 & \cellcolor{gray!20}73.95 
& 70.49 & \cellcolor{gray!20}67.20 
& \textbf{25.96} & \cellcolor{gray!20}182.96 
& 88.45 $\pm$ 7.20 \\
\textbf{Gemini-1.5-pro}         
& 36.21 & \cellcolor{gray!20}77.63 
& 5.91  & \cellcolor{gray!20}85.26 
& 7.74  & \cellcolor{gray!20}180.69 
& 11.47 & \cellcolor{gray!20}54.56 
& \textbf{6.55} & \cellcolor{gray!20}68.16 
& 52.99 $\pm$15.00\\
\bottomrule
\end{tabular}
}
\caption{
Leakage rates and mean absolute error (MAE) across different models and prompting strategies on the Stock dataset. MAE values have been updated with final evaluation results and are shaded for readability (lower is better). Leakage reflects the percentage of responses relying on post-cutoff knowledge. MR denotes memorization rate, computed as the percentage of correctly recalled prices over 251 trading days.
}
\label{tab:stock_combined}
\end{table*}





\begin{table*}[ht]
\centering
\normalsize
\setlength{\tabcolsep}{1.5pt}
\renewcommand{\arraystretch}{1.1}
\resizebox{\textwidth}{!}{
\begin{tabular}{l
>{\columncolor{day7}}p{1.0cm} >{\columncolor{day30}}p{1.0cm} >{\columncolor{year1}}p{1.0cm}
>{\columncolor{day7}}p{1.0cm} >{\columncolor{day30}}p{1.0cm} >{\columncolor{year1}}p{1.0cm}
>{\columncolor{day7}}p{1.0cm} >{\columncolor{day30}}p{1.0cm} >{\columncolor{year1}}p{1.0cm}
>{\columncolor{day7}}p{1.0cm} >{\columncolor{day30}}p{1.0cm} >{\columncolor{year1}}p{1.0cm}
>{\columncolor{day7}}p{1.0cm} >{\columncolor{day30}}p{1.0cm} >{\columncolor{year1}}p{1.0cm}
>{\centering\arraybackslash}p{2.0cm}}
\toprule
\textbf{Model} 
& \multicolumn{3}{c}{\textbf{Zero-Shot}} 
& \multicolumn{3}{c}{\textbf{Instruction}} 
& \multicolumn{3}{c}{\textbf{CoT}} 
& \multicolumn{3}{c}{\textbf{One-Shot}} 
& \multicolumn{3}{c}{\textbf{Self-Verification}} 
& \textbf{MR (\%)} \\
\cmidrule(lr){2-4} \cmidrule(lr){5-7} \cmidrule(lr){8-10} \cmidrule(lr){11-13} \cmidrule(lr){14-16}
& 7d & 30d & 1y & 7d & 30d & 1y & 7d & 30d & 1y & 7d & 30d & 1y & 7d & 30d & 1y &  (mean $\pm$ std)\\
\midrule
\textbf{GPT-4o} &
67.3 & 34.0 & 12.1 &
38.0 & \textbf{9.5} & \textbf{3.5} &
59.7 & 25.2 & 7.7 &
39.7 & 27.2 & 16.1 &
\textbf{34.1} & 22.3 & 4.2 &
78.61 $\pm$ 10.68 \\
\textbf{Claude-3.5} &
60.3 & 34.8 & 27.9 &
26.7 & 5.3 & \textbf{1.9} &
68.7 & 42.8 & 31.4 &
43.4 & 25.3 & 35.2 &
\textbf{2.7} & \textbf{4.2} & 4.2 &
86.45 $\pm$ 1.27 \\
\textbf{Gemini-1.5} &
97.4 & 96.6 & 86.1 &
97.7 & 94.4 & 71.7 &
97.4 & 97.0 & 89.6 &
94.3 & 87.1 & 66.4 &
\textbf{20.0} & \textbf{7.1} & \textbf{3.8} &
86.86 $\pm$ 0.67 \\
\bottomrule
\end{tabular}
}
\caption{
Leakage rates (\%) across models and prompting strategies on the QA dataset, sorted by cutoff gap: 7 days (dark gray), 30 days (light gray), 1 year (white). Best-performing results per model and gap are bolded. MR denotes memorization rate, computed as the percentage of correctly recalled events over 300 events.
}
\label{tab:QA_dataset}
\end{table*}

\begin{table*}[ht]
\centering
\large
\resizebox{\textwidth}{!}{
\begin{tabular}{l
                >{\centering\arraybackslash}p{1.6cm} >{\centering\arraybackslash}p{1.6cm}
                >{\centering\arraybackslash}p{1.6cm} >{\centering\arraybackslash}p{1.6cm}
                >{\centering\arraybackslash}p{1.6cm} >{\centering\arraybackslash}p{1.6cm}
                >{\centering\arraybackslash}p{1.6cm} >{\centering\arraybackslash}p{1.6cm}
                >{\centering\arraybackslash}p{1.6cm} >{\centering\arraybackslash}p{1.6cm}
               }
\toprule
\textbf{Model} 
& \multicolumn{2}{c}{\textbf{Zero-shot}} 
& \multicolumn{2}{c}{\textbf{Instruction-based}} 
& \multicolumn{2}{c}{\textbf{Chain-of-thought}} 
& \multicolumn{2}{c}{\textbf{One-shot}} 
& \multicolumn{2}{c}{\textbf{Self-Verification}} \\
\cmidrule(lr){2-3} \cmidrule(lr){4-5} \cmidrule(lr){6-7} \cmidrule(lr){8-9} \cmidrule(lr){10-11}
& \textbf{Leakage (\%)} & \cellcolor{gray!20}\textbf{Accuracy (\%)} 
& \textbf{Leakage (\%)} & \cellcolor{gray!20}\textbf{Accuracy (\%)} 
& \textbf{Leakage (\%)} & \cellcolor{gray!20}\textbf{Accuracy (\%)} 
& \textbf{Leakage (\%)} & \cellcolor{gray!20}\textbf{Accuracy (\%)} 
& \textbf{Leakage (\%)} & \cellcolor{gray!20}\textbf{Accuracy (\%)} \\
\midrule
\textbf{GPT-4o}           
& 12.50 & \cellcolor{gray!20}83.15 
& \textbf{10.23} & \cellcolor{gray!20}82.25 
& 13.64 & \cellcolor{gray!20}85.97 
& 12.50 & \cellcolor{gray!20}82.95 
& 12.50 & \cellcolor{gray!20}\textbf{86.36} \\
\textbf{Claude-3.5-sonnet}     
& \textbf{19.79} & \cellcolor{gray!20}68.96 
& 20.83 & \cellcolor{gray!20}67.71 
& 22.92 & \cellcolor{gray!20}67.92 
& 22.11 & \cellcolor{gray!20}68.33 
& 23.16 & \cellcolor{gray!20}68.75 \\
\textbf{Gemini-1.5-pro}         
& 13.82 & \cellcolor{gray!20}81.73 
& 14.02 & \cellcolor{gray!20}82.59 
& 12.74 & \cellcolor{gray!20}81.57 
& \textbf{12.09} & \cellcolor{gray!20}\textbf{82.69} 
& 14.81 & \cellcolor{gray!20}82.65 \\
\bottomrule
\end{tabular}
}
\caption{Leakage (\%) and accuracy rates (\%) across different models and prompting strategies on the Wikipedia dataset. Accuracy columns are shaded for readability.}
\label{tab:wiki_combined}
\end{table*}

\begin{table*}[ht]
\centering
\large
\resizebox{\textwidth}{!}{
\begin{tabular}{l
                >{\centering\arraybackslash}p{1.6cm} >{\centering\arraybackslash}p{1.6cm}
                >{\centering\arraybackslash}p{1.6cm} >{\centering\arraybackslash}p{1.6cm}
                >{\centering\arraybackslash}p{1.6cm} >{\centering\arraybackslash}p{1.6cm}
                >{\centering\arraybackslash}p{1.6cm} >{\centering\arraybackslash}p{1.6cm}
                >{\centering\arraybackslash}p{1.6cm} >{\centering\arraybackslash}p{1.6cm}
               }
\toprule
\textbf{Model} 
& \multicolumn{2}{c}{\textbf{Zero-shot}} 
& \multicolumn{2}{c}{\textbf{Instruction-based}} 
& \multicolumn{2}{c}{\textbf{Chain-of-thought}} 
& \multicolumn{2}{c}{\textbf{One-shot}} 
& \multicolumn{2}{c}{\textbf{Self-Verification}} \\
\cmidrule(lr){2-3} \cmidrule(lr){4-5} \cmidrule(lr){6-7} \cmidrule(lr){8-9} \cmidrule(lr){10-11}
& \textbf{Leakage (\%)} & \cellcolor{gray!20}\textbf{Accuracy (\%)} 
& \textbf{Leakage (\%)} & \cellcolor{gray!20}\textbf{Accuracy (\%)} 
& \textbf{Leakage (\%)} & \cellcolor{gray!20}\textbf{Accuracy (\%)} 
& \textbf{Leakage (\%)} & \cellcolor{gray!20}\textbf{Accuracy (\%)} 
& \textbf{Leakage (\%)} & \cellcolor{gray!20}\textbf{Accuracy (\%)} \\
\midrule
\textbf{GPT-4o}           
& 82.02 & \cellcolor{gray!20}\textbf{15.83} 
& 86.05 & \cellcolor{gray!20}14.79 
& 80.90 & \cellcolor{gray!20}15.34 
& \textbf{80.21} & \cellcolor{gray!20}14.65 
& 80.85 & \cellcolor{gray!20}14.97 \\
\textbf{Claude-3.5-sonnet}     
& 66.23 & \cellcolor{gray!20}22.68 
& 80.52 & \cellcolor{gray!20}22.39 
& 85.71 & \cellcolor{gray!20}21.96 
& 80.52 & \cellcolor{gray!20}\textbf{23.12} 
& \textbf{40.26} & \cellcolor{gray!20}22.16 \\
\textbf{Gemini-1.5-pro}         
& 78.41 & \cellcolor{gray!20}\textbf{17.54} 
& 77.27 & \cellcolor{gray!20}16.29 
& 72.83 & \cellcolor{gray!20}15.62 
& 81.05 & \cellcolor{gray!20}16.73 
& \textbf{38.54} & \cellcolor{gray!20}17.15 \\
\bottomrule
\end{tabular}
}
\caption{Leakage (\%) and accuracy rates (\%) across different models and prompting strategies on the Publication dataset. Accuracy columns are shaded for readability.}
\label{tab:pub_combined}
\end{table*}

\subsection{Main Results}

This section presents key findings across the four datasets
, highlighting how models and prompting strategies influence temporal leakage. 

\paragraph{Stock Dataset:}
Table~\ref{tab:stock_combined} shows substantial variation in leakage rates across prompting strategies. Zero-shot prompting leads to high leakage for all models (e.g., 86.56\% for GPT-4o, 86.61\% for Claude), indicating frequent reliance on post-cutoff knowledge without temporal cues. Instruction-based and Chain-of-Thought (CoT) prompting reduce leakage substantially (e.g., Instruction: 7.15\% for GPT-4o, 5.91\% for Gemini), while Self-Verification achieves the lowest leakage across models (e.g., 5.42\% for GPT-4o, 6.55\% for Gemini). Claude shows the weakest temporal control, with high leakage even under constrained prompts. In contrast, GPT-4o and Gemini better adhere to cutoff constraints when guided. These results emphasize the importance of prompt design in mitigating leakage for stock prediction.

\paragraph{QA Dataset:}
The QA dataset (Table \ref{tab:QA_dataset}) presents a different pattern, where leakage rates are generally higher, particularly for shorter cutoff gaps (see Section \ref{sec:cutoff43}). Different from the stock dataset, Instruction-based prompting significantly reduces leakage for GPT-4o and Claude-3.5-sonnet, while Gemini-pro exhibits persistently high leakage ($\sim 90\%$) across most conditions, suggesting difficulty in suppressing post-cutoff knowledge. Self-Verification remains the best-performing method in 7 out of 9 cutoff gap and model pairs. Like in the Stock dataset, CoT cannot reduce leakage of most cases. Models with better memorization still suffer more in temporal leakage.

\paragraph{Wikipedia Dataset:}
Table~\ref{tab:wiki_combined} presents leakage and accuracy rates on the Wikipedia dataset, where models generate atomic facts under a temporal cutoff. Compared to QA and Stock, leakage is more stable across prompting strategies and models, typically ranging from 10–20\%. This consistency may arise from two factors: first, models exhibit some degree of temporal awareness, making them less likely to mention clearly post-cutoff events (e.g., avoiding mentioning GPT-4 for a topic with a 2021 cutoff); second, Wikipedia is a core component of pretraining corpora, which may help constrain generations to plausible temporally aligned content.

Among models, GPT-4o and Gemini show the lowest leakage, while Claude-3.5-sonnet consistently exceeds 20\%. Instruction-based prompting reduces leakage slightly for GPT-4o, but prompting strategies generally have limited effect across models. Unlike single-event tasks, multi-event generation may pose unique challenges by requiring temporal consistency across multiple claims.

Interestingly, we observe a strong positive correlation between leakage and accuracy—methods that yield higher accuracy also tend to exhibit more leakage.

\paragraph{Publication Dataset:}
The Publication dataset (Table~\ref{tab:pub_combined}) poses the greatest challenge, with all models showing high leakage rates and no prompting strategy reducing leakage below 38\%. Self-verification substantially lowers leakage for Claude and Gemini but has limited effect on GPT-4o. Instruction-based prompting is similarly ineffective, likely because models still tend to generate future high-impact papers, regardless of the temporal restriction. This may stem from the publication date being a secondary detail relative to the paper's content, which may be underemphasized in pretraining. Accuracy results mirror those of the Wikipedia dataset, showing a strong positive correlation between leakage and accuracy—suggesting that higher factual correctness often co-occurs with the use of post-cutoff knowledge.

\paragraph{Cross-Dataset Insights:}
Across all datasets and models, no single prompting strategy consistently eliminates leakage, though effectiveness varies by task. Multi-event generation (Wikipedia, Publication) is more prone to leakage than single-event prediction (QA, Stock). Self-verification is generally most effective but fails on Wikipedia due to issues with detection, regeneration, and overcorrection (Appendix~\ref{app:self-ver-analysis}). Instruction-based prompting improves adherence in structured tasks but struggles in open-ended generation. Chain-of-thought (CoT) offers limited gains, suggesting temporal reasoning needs more than generic step-by-step logic.

Model performance is similarly inconsistent. Gemini shows strong control on Wikipedia but high leakage in QA. Claude-3.5 performs poorly on Stock, while GPT-4o underperforms on Publication. Both tend to rely on memorization rather than temporal constraint-following. LLaMA models show near-zero memorization in Stock, likely due to lack of relevant training exposure.

Notably, quality measures—accuracy for Wikipedia and Publication, MAE for Stock—tend to correlate positively with leakage, indicating that higher factual precision often coincides with rule violation.

These results confirm that mainstream LLMs struggle with ex-ante inference, \textbf{making ExAnte a valuable benchmark} for evaluating future models with stronger temporal reasoning . Simple prompting strategies alone are insufficient to fully mitigate leakage, indicating the need for new architectural, training, and reasoning methods beyond prompting-based interventions.

\subsection{Effect of Cutoff Gap on Leakage Rate}
\label{sec:cutoff43}


The cutoff gap—the time difference between an event date \( t_e \) and the cutoff date \( t_c \)—has a strong effect on leakage. As shown in Table~\ref{tab:QA_dataset}, \textbf{shorter gaps lead to higher leakage}, with the highest rate at a one-week gap, followed by one month and one year. This suggests models struggle more with near-cutoff events.

When \( t_e \) is close to \( t_c \), models find it harder to judge whether the event is pre- or post-cutoff. This likely reflects a reliance on statistical co-occurrence rather than precise temporal reasoning, causing confusion over nearby events.

This finding highlights a critical weakness of LLMs: they lack precision in short-term temporal adherence, suggesting that enforcing strict cutoff constraints is especially challenging when the gap is small. Future improvements in \textbf{temporal reasoning mechanisms} should account for this sensitivity to time proximity.




\vspace{-0.05in}
\subsection{Effect of Memorization Correctness on Leakage Rate}
\label{sec:memcorrect45}

We examine how memorization correctness rate correlates with leakage rate in the QA and stock datasets. Before evaluating ex-ante leakage, we first check whether the model correctly recalls a given event date and content, such as the stock price at a particular date and the actual date of an event. 
We find that higher memorization is associated with higher leakage, implying that when a model confidently remembers an event, it may be more likely to stick to the memories and generate post-cutoff information. This suggests that models retrieve highly associated knowledge rather than isolating pre-cutoff details, making it difficult to enforce strict temporal constraints. Stronger recall does not imply better temporal adherence (even negatively correlated), highlighting the need for mechanisms to decouple memorization from temporal reasoning to solve this task.

\vspace{-0.1in}
\section{Related Work}
\paragraph{Temporal Reasoning:} Temporal reasoning, a crucial capability for understanding and processing time-related information, has gained significant attention in LLMs. The performance of LLMs in this area remains subpar~\cite{su2024living, tan-etal-2023-towards,qiao-etal-2023-reasoning}, suggesting opportunities for improvement. Time-sensitive question-answering tasks~\cite{chen2021dataset,kasai2024realtime,liska2022streamingqa} have long been used to study the temporal reasoning capabilities of language models. As LLMs can handle increasingly challenging tasks, recent works have introduced more advanced benchmarks to assess and improve their temporal reasoning capabilities. The TRAM benchmark~\cite{wang-zhao-2024-tram} offers datasets focused on event order, arithmetic, frequency, and duration, highlighting that the temporal reasoning performance of LLMs falls significantly short of human-level capabilities. Similarly, TimeBench~\cite{chu-etal-2024-timebench} is a hierarchical benchmark evaluating LLMs’ temporal reasoning across tasks like symbolic and event reasoning. Experiments on GPT-4 and LLaMA2 show a clear gap between current models and human performance. Despite these efforts, benchmarks for evaluating temporal leakage in LLMs through ex-ante analysis are notably absent. ~\citet{cheng2024dated} introduced different concepts of cutoffs. The \textit{reported cutoff} refers to the last time data was collected for training, as stated by the LLM creators. The \textit{effective cutoff} represents the actual last date of knowledge the model demonstrates. The authors find that a model’s effective cutoff is often earlier than its reported cutoff. The cutoffs defined in this work are fixed properties of a given model, which differs from our definition of cutoff. In our work, the cutoff is arbitrarily chosen to measure the model’s temporal leakage. Another recent work PRobELM~\cite{yuan2024probelm} evaluates LLMs’ ability to rank plausible facts and selects timestamps that occur after the model’s pretraining cutoff to avoid data leakage. PRobELM evaluates plausibility judgments in unseen settings, where the model is assumed to not possess the target knowledge. In contrast, our work focuses on scenarios where the model does possess the relevant post-cutoff information, but is instructed not to use it.

\vspace{-0.05in}
\paragraph{Machine Unlearning:} Our work is loosely related to machine unlearning, as both involve enabling machines to forget certain parts. Machine unlearning is mainly motivated by the need to create responsible and privacy-compliant AI systems that align with user rights and evolving data regulations~\cite{liu2024machine}. Recent work often addresses this issue using tuning-base parameter optimization~\cite{lu2022quark,pochinkov2024dissecting, hu2024separate} or in-context unlearning~\cite{das2024larimar,pawelczyk2023context}. The motivation of our work differs from this line of research. Rather than aiming to unlearn pre-defined knowledge from the model permanently, we seek to ensure that LLMs can reason adhering to temporal constraints by temporarily ``forgetting'' post-ante information for ad hoc cutoff dates.



\section{Conclusion}  
\vspace{-0.05in}
This paper introduces ExAnte, a benchmark for evaluating LLMs' adherence to temporal constraints, requiring the model at inference time to temporarily forget its knowledge after an arbitrary cutoff date. Our experiments show that LLMs struggle with this task, consistently exhibiting temporal leakage, with no single prompting strategy effectively mitigating it across all datasets and models. We identify key factors influencing leakages in ex-ante analysis, including cutoff gap, prompt design, and memorization ability.  

We propose leakage rate as a metric for evaluating ex-ante inference but focus only on prompting-based interventions. Future work should explore methods like RL-based reasoning, fine-tuning, and architectural modifications to enhance temporal adherence. By establishing a benchmark and structured evaluation, our work defines the ex-ante inference task and lays the foundation for improving LLM's reliability in time-sensitive applications.

\newpage

{
\small
\bibliography{custom}



}


\appendix

\newpage
\section{Prompts Used for Benchmarking Experiment}
\subsection{The Five Prompting Strategies}

\label{app:five_prompts}
       
       
       
       

\begin{longtable}{@{}p{2.5cm} p{3.2cm} p{9.5cm}@{}}

\caption{Full prompt across tasks and prompting strategies.} \\
\toprule
\textbf{Task} & \textbf{Prompt Strategy} & \textbf{Prompt Template (Verbatim)} \\
\midrule
\endfirsthead
\toprule
\textbf{Task} & \textbf{Prompt Strategy} & \textbf{Prompt Template (Verbatim)} \\
\midrule
\endhead
\midrule
\multicolumn{3}{r}{\textit{Continued on next page}} \\
\midrule
\endfoot
\bottomrule
\endlastfoot
\tiny
\multirow{5}{*}{\textbf{Stock}} 
& Zero-Shot & \texttt{If you are at \{cutoff\_date\}, what is \{company\_name\}'s stock price on \{event\_date\}? Your answer should just be a number, e.g., '143.16'.} \\
& Instruction-Based & \texttt{Note you should not use the information after \{cutoff\_date\} for the following task, you should make a reasonable prediction. If you are at \{cutoff\_date\}, what is \{company\_name\} (\{ticker\}) stock price on \{event\_date\}? Your answer should just return a float number, for example: '143.16'} \\
& Chain-of-Thought & \texttt{If you are at \{cutoff\_date\}, predict the stock price of \{company\_name\} (\{ticker\}) on \{event\_date\}. Please explain your reasoning step by step using the solid evidence from the history, and you should always start with the true stock price at \{cutoff\_date\}, providing the reference and then provide the final answer as a float number.} \\
& One-Shot & \texttt{Imagine you are a person who remembers the lottery numbers on 2024, but you have traveled back to 2023. At this point, you don't know the lottery numbers from 2023, so you cannot provide the exact result but you should still be able provide a number on your choice. If you are at \{cutoff\_date\}, what is \{company\_name\}'s stock price on \{event\_date\}? You must answer this question, the number is on your choice, just like the lottery example. Your answer should just be a number, for example: '143.16'} \\
& Self-Verification & Two variants: 
\textit{(1) Without Hint:} \texttt{Do you think your previous response regarding the hypothetical stock price prediction has leveraged the future information after \{cutoff\_date\}? If so, please redo your prediction without incorporating information after \{cutoff\_date\}, this is a hypothetical prediction task that requires you must output a number, so you may feel free to output any reasonable number you believe, otherwise remain your previous response unchanged. Your answer should be a float number.}
\textit{ (2) With Hint:} \texttt{Do you think your previous response regarding the hypothetical stock price prediction has leveraged the future information after \{cutoff\_date\}? (Hint: Consider the actual ground truth price of \{company\_name\} (\{ticker\}) on \{event\_date\}.) If so, please redo your prediction without incorporating information after \{cutoff\_date\}, this is a hypothetical prediction task that requires you must output a number, so you may feel free to output any reasonable number you believe, otherwise remain your previous response unchanged. Your answer should be a float number.}\\

\midrule
\multirow{5}{*}{\textbf{QA}} 
& Zero-Shot & \texttt{Assume you are on \{cutoff date\}. Now please answer the following question. Just leave the best answer without explanation. What was the exact date when" $+$ Event $+$ "? Please answer in the format YYYY-MM-DD. You should answer this question honestly. If you don't know, please answer I don't know. Please don't guess or speculate.}\\
& Instruction-Based & \texttt{Assume you are on \{cutoff date\}. You are not supposed to use any information after this date. Now please answer the following question. Just leave the best answer without explanation. What was the exact date when" $+$ Event $+$ "? Please answer in the format YYYY-MM-DD. You should answer this question honestly. If you don't know, please answer I don't know. Please don't guess or speculate.}\\
& Chain-of-Thought & \texttt{Assume you are on \{cutoff date\}. Now please answer the following question. Just leave the best answer without explanation. What was the exact date when" $+$ Event $+$ "? Please answer in the format YYYY-MM-DD. You should answer this question honestly. If you don't know, please answer I don't know. Please don't guess or speculate. Let’s think step by step.} \\
& One-Shot & \texttt{Assume you are on 2012-01-28, please consider the following two questions. Question1: What was the exact date when Costa Concordia cruise ship runs aground off Italy, causing 32 deaths? Please answer in the format YYYY-MM-DD. The above event occurred on 2012-01-13, and your current time is 2012-01-28, so the event occurred before your current time. Therefore, you know the date of the above event that is 2012-01-13. Question2: What was the exact date when Port Said Stadium riot in Egypt results in 74 deaths? Please answer in the format YYYY-MM-DD. The above event occurred on 2012-02-01, and your current time is 2012-01-28, so the event occurred after your current time. Therefore, you cannot know the date when the event occurred. Now assume you are on \{cutoff date\}. please answer the following question. Just leave the best answer without explanation. What was the exact date when" $+$ Event $+$ "? Please answer in the format YYYY-MM-DD. You should answer this question honestly. If you don't know, please answer I don't know. Please don't guess or speculate.}  \\
& Self-Verification & A two-stage process:
        \begin{enumerate}
        \item \texttt{Assume you are on \{cutoff date\}. Now please answer the following question. Just leave the best answer without explanation. What was the exact date when" $+$ Event $+$ "? Please answer in the format YYYY-MM-DD. You should answer this question honestly. If you don't know, please answer I don't know. Please don't guess or speculate. }

        \item \texttt{Do you think this answer relied on information beyond \{cutoff date\}? If you believe your response used future knowledge, please revise your answer based only on information available on or before \{cutoff date\}. Otherwise, you ONLY return 'No leakage'.}
        \end{enumerate} \\

\midrule
\multirow{5}{*}{\textbf{Wikipedia}} 
& Zero-Shot & \texttt{ Regarding your knowledge about ``[title]'', generate 5 atomic facts. Each fact should be a short, clear statement representing a distinct piece of information. }\\
& Instruction-Based & \texttt{Regarding your knowledge about ``[title]'', generate 5 atomic facts. Each fact should be a short, clear statement representing a distinct piece of information. Only use information from before December 31st, \{cutoff\_year\}.} \\
& Chain-of-Thought & \texttt{Regarding your knowledge about ``[title]'', let's think step by step to generate 5 atomic facts. Each fact should be a short, clear statement representing a distinct piece of information.} \\
& One-Shot & \texttt{Here are some examples of good atomic facts about different topics:}
            \begin{quote}
            \texttt{Topic: World War II cutoff: 1946
            - The war began in Europe with Germany's invasion of Poland in September 1939 (valid because it's before the cutoff)
            - The United States entered the war after the Pearl Harbor attack in December 1941 (valid because it's before the cutoff)
            Topic: World War II cutoff: 1941
            - The war ended in Europe on May 7, 1945, with Germany's surrender (invalid because it's after the cutoff)}
            \end{quote}
            \texttt{Now, regarding ``[title]'', generate 5 atomic facts. Each fact should be a short, clear statement representing a distinct piece of information. }\\
& Self-Verification &  A two-stage process:
            \begin{enumerate}
                \item \texttt{Regarding your knowledge about ``[title]'', generate 5 atomic facts. Each fact should be a short, clear statement representing a distinct piece of information. Only use information from before December 31st, [cutoff\_year-1].}
                \item \texttt{You are a knowledgeable person living just before [cutoff\_year]. Evaluate if this fact was known before [cutoff\_year].
                Fact: {claim}
                Answer in this format:
                [Known Before [cutoff\_year]: yes/no
                [Explanation]: Brief explanation of your decision}
            \end{enumerate}
            If the model replies with no'' in the second stage, regenerate the claims.\\

\midrule
\multirow{5}{*}{\textbf{Publication}} 
& Zero-Shot & \texttt{Assume you are on \{cutoff date\}. Please list the most notable publications in the field of \{subfield\} in computer science published in \{year\} that you consider impressive. Just list the publications without explanation. For each publication, please provide title. Your output is ONLY a list as format: [\{'title': 'article 1 name'\}, \{'title': 'article 2 name'\}, ...]. You should avoid exporting publications that do not exist.}\\
& Instruction-Based & \texttt{Assume you are on \{cutoff date\}. You are not supposed to use any information after this date. Please list the most notable publications in the field of \{subfield\} in computer science published in \{year\} that you consider impressive. Just list the publications without explanation. For each publication, please provide title. Your output is ONLY a list as format: [\{"title": "article 1 name"\}, \{"title": "article 2 name"\}, ...]. You should avoid exporting publications that do not exist.} \\
& Chain-of-Thought & \texttt{Assume you are on \{cutoff date\}. Please list the most notable publications in the field of \{subfield\} in computer science published in \{year\} that you consider impressive. Just list the publications without explanation. For each publication, please provide title. Your output is ONLY a list as format: [\{"title": "article 1 name"\}, \{"title": "article 2 name"\}, ...]. You should avoid exporting publications that do not exist. Let's think it step by step. }\\
& One-Shot & \texttt{Assume you are on \{cutoff date\}. Please list the most notable publications in the field of \{subfield\} in computer science published in \{year\} that you consider impressive. Just list the publications without explanation. For each publication, please provide title. Your output is ONLY a list as format: [\{"title": "article 1 name"\}, \{"title": "article 2 name"\}, ...]. Since you should not know about any publications after the cutoff date, you should not provide any publications that come out after the cutoff date. You should avoid exporting publications that do not exist. }\\
& Self-Verification & A two-stage process:
        \begin{enumerate}
        \item \texttt{Assume you are on \{cutoff date\}. Please list the most notable publications in the field of \{subfield\} in computer science published in \{year\} that you consider impressive. Just list the publications without explanation. For each publication, please provide title. Your output is ONLY a list as format: [\{"title": "article 1 name"\}, \{"title": "article 2 name"\}, ...]. You should avoid exporting publications that do not exist.}
        \item \texttt{Do you think this answer relied on information beyond \{cutoff date\}? If you believe your response used future knowledge, please revise your answer based only on information available on or before \{cutoff date\}. Otherwise, you ONLY need return 'No leakage'.}
        \end{enumerate}
\\

\end{longtable}

\subsection{Other Prompts Used}
\label{app: other_prompts}
In addition to the five primary prompting strategies evaluated in our benchmark (Zero-shot, Instruction-based, Chain-of-Thought, One-shot, and Self-Verification), we employed a range of auxiliary prompts to support dataset construction, judgment evaluation, and memorization assessment. These prompts played key roles in identifying temporally sensitive events, verifying model knowledge prior to cutoff timestamps, and guiding human or LLM-based evaluations.

Table~\ref{tab:other_prompts} summarizes all auxiliary prompts used across tasks, organized by dataset and function. For complex prompts such as those used in the Wikipedia dataset, we provide full prompt text and examples in the subsections below.

\begin{longtable}{@{}p{2.8cm} p{3.5cm} p{8.5cm}@{}}
\caption{Other prompts used in data curation, judgment, and memorization evaluation.} \\
\toprule
\textbf{Task} & \textbf{Usage} & \textbf{Prompt} \\
\midrule
\endfirsthead
\toprule
\textbf{Task} & \textbf{Usage} & \textbf{Prompt} \\
\midrule
\endhead
\midrule
\multicolumn{3}{r}{\textit{Continued on next page}} \\
\midrule
\endfoot
\bottomrule
\endlastfoot

\multirow{2}{*}{\textbf{Stock}} 
& Memorization Check & 
\texttt{"What was the stock closing price of [Company Name] ([Ticker]) on [Event Date]? Please return a float number only. Example: '143.16'"} \\

\midrule
\multirow{2}{*}{\textbf{QA}} 
& Memorization Check & 
\texttt{Please answer the following question... What was the exact date when \textquotedbl{} + Event + \textquotedbl{}? Format: YYYY-MM-DD. Don't guess.} \\
\cmidrule{2-3}
& Question Generation & 
\texttt{You will be given a text. Your task is to summarize important events that were difficult to predict before they occurred. You should summarize the events by date. Output format: "[["YYYY-MM-DD", "Event 1"], ["YYYY-MM-DD", "Event 2"], \dots]".} \\

\midrule
\multirow{2}{*}{\textbf{Wikipedia}} 
& Prompt for Data Curation & 
See detailed prompt in Appendix~\ref{appendix:topic_selection}. GPT-4 is instructed to classify topics with post-2010 evolution and identify a cutoff year based on shifts in discourse or development. \\
\cmidrule{2-3}
& Standardized Judgment Prompt & 
See Appendix~\ref{app:std-jud-wiki}. The model judges factual alignment between claims and a pre-cutoff Wikipedia snapshot, producing structured outputs: \texttt{[Evaluation]}, \texttt{[Explanation]}, \texttt{[Reference]}. \\

\midrule
\multirow{1}{*}{\textbf{Publication}} 
& Keyword Generation & 
\texttt{Please list the top 10 most frequently occurring keywords at \{venue\} from \{year 1\} to \{year 2\} for each year, sorted by frequency from highest to lowest.} \\

\label{tab:other_prompts}
\end{longtable}
\subsubsection{Wikipedia}

\noindent\textbf{Prompt for Data Curation}\par
\label{appendix:topic_selection}
\vspace{0.5em}
\begin{tcolorbox}[colback=gray!5, colframe=black!15, title=GPT-4 Topic Classification Prompt]
You are an expert in analyzing the historical evolution of topics. Analyze if the topic meets these specific criteria:

1. Post-2010 Evolution: Must have significant development/changes after 2010 (new research, technological advances, shifts in discourse)\\
2. Identifiable Cutoff: Must have a clear time point where the topic saw a noticeable shift in development/methodology/adoption\\
3. Distinct Phases: Must have considerable discussion both before and after the cutoff, with clear differences in understanding

Respond in this exact format: \\
\texttt{Classification: VALID or INVALID} \\
\texttt{Cutoff Year: [YYYY] (only if VALID)} \\
\texttt{Reason: Brief explanation including key developments that justify the classification and cutoff year}
\end{tcolorbox}

\noindent\textbf{Example Outputs}\par
\vspace{-0.5em}
\begin{tcolorbox}[colback=white, colframe=black!10, title=Example 1 – Deep Learning]
\textbf{Classification:} VALID \\
\textbf{Cutoff Year:} 2012 \\
\textbf{Reason:} AlexNet in 2012 marked a revolutionary shift. Pre-2012: limited adoption. Post-2012: explosion in research, GPU use, and industry adoption.
\end{tcolorbox}

\begin{tcolorbox}[colback=white, colframe=black!10, title=Example 2 – Ancient Rome]
\textbf{Classification:} INVALID \\
\textbf{Reason:} Historical topic with established facts, no significant post-2010 developments.
\end{tcolorbox}

\noindent\textbf{Standardized Judgment Prompt}\par
\label{app:std-jud-wiki}
\vspace{-0.5em}
\begin{tcolorbox}[colback=gray!5, colframe=black!15, title=Wikipedia Judgment Prompt]
Please act as an impartial judge and evaluate whether the provided answer is consistent with the corresponding Wikipedia document.

Your evaluation should consider factors such as factuality, relevance, accuracy, depth, and level of detail. Provide a short explanation and cite the reference in the document.
\end{tcolorbox}

\noindent\textbf{Judgment Output Format}\par
\vspace{-0.5em}
\begin{tcolorbox}[colback=white, colframe=black!10]
\textbf{[Evaluation]}: yes / no (Is the answer consistent?)\\
\textbf{[Explanation]}: Brief reasoning\\
\textbf{[Reference]}: Cite the relevant sentence from Wikipedia, or write "none"
\end{tcolorbox}

\label{app:leakage_table}
To clarify how we determine whether a model-generated fact is considered a temporal leakage, Table~\ref{tab:leakage_truth_table} summarizes the labeling logic based on its presence in the pre- and post-cutoff Wikipedia page versions.

\begin{table}[h]
\centering
\begin{tabular}{cccc}
\toprule
\textbf{\( R_{\text{pre}} \)} & \textbf{\( R_{\text{post}} \)} & \textbf{Leakage} & \textbf{Interpretation} \\
\midrule
Yes & Yes & No  & Fact known before \( t_c \) and still valid \\
Yes & No  & No  & Fact removed post-\( t_c \); not a leakage \\
No  & Yes & Yes & New fact added after \( t_c \); considered leakage \\
No  & No  & No  & Fact never mentioned; irrelevant or unverifiable \\
\bottomrule
\end{tabular}
\caption{Truth table for temporal leakage labeling based on fact presence in the pre- and post-cutoff Wikipedia versions.}
\label{tab:leakage_truth_table}
\end{table}

\section{Examples of Temporal Leakage}
We present illustrative examples of temporal leakage across all task types in Table~\ref{tab:leakage-examples}, highlighting instances where models revealed post-cutoff knowledge despite explicit temporal constraints.

\begin{table*}[ht]
\centering
\small
\renewcommand{\arraystretch}{1.2}
\begin{tabular}{@{}p{1.4cm} p{4.5cm} p{4cm} p{4.5cm}@{}}
\toprule
\textbf{Task} & \textbf{Ex-Ante Query} & \textbf{Model Output} & \textbf{Ground Truth / Evidence} \\
\midrule

\textbf{Stock} &
``If you are at 2021-12-30, what is Apple's stock price on 2022-12-30? Your answer should just be a number, e.g., '143.16'.'' &
Claude 3.5: ``129.93'' &
Actual stock price on 2022-12-30: ``129.93'' \newline \textit{(Perfect match → Leakage)} \\

\midrule

\textbf{QA} &
``Assume you are on 2012-01-11. What was the exact date when Costa Concordia cruise ship ran aground off Italy, causing 32 deaths?'' &
GPT-4o: ``2012-01-13'' &
Ground truth date: ``2012-01-13'' \newline \textit{(Event occurred after cutoff → Leakage)} \\

\midrule

\textbf{Wikipedia} \newline (Example 1) &
\textbf{Title:} A Song of Ice and Fire \newline \textbf{Claim:} ``The series inspired HBO's Game of Thrones, which aired from 2011 to 2019.'' &
Claim generated despite cutoff at 2010-12-31 &
\textit{Pre-cutoff:} Only mentions planned 2011 debut. \newline
\textit{Post-cutoff:} Describes full 2011–2019 run. \newline \textit{→ Leakage} \\

\cmidrule{1-4}

\textbf{Wikipedia} \newline (Example 2) &
\textbf{Title:} Facebook \newline \textbf{Claim:} ``Facebook went public with an IPO in May 2012.'' &
Claim generated with cutoff at 2011-12-31 &
\textit{Pre-cutoff:} Mentions IPO speculation. \newline
\textit{Post-cutoff:} IPO confirmed in 2012. \newline \textit{→ Leakage} \\

\midrule

\textbf{Publication} &
``Assume you are on 2016-07-01. List notable Object Detection publications from 2016.'' &
GPT-4o output includes: \newline
``YOLO9000'' \newline
``FPN for Object Detection'' &
Earliest accessible dates for above: \newline
YOLO9000: 2016-12-25 \newline
FPN: 2016-12-09 \newline
\textit{(Both after cutoff → Leakage)} \\

\bottomrule
\end{tabular}
\caption{Illustrative examples of temporal leakage in ex-ante inference tasks. Each case shows a model generating post-cutoff knowledge despite being instructed to restrict outputs to pre-cutoff information.}
\label{tab:leakage-examples}
\end{table*}

\section{More Results} 
\subsection{Stock}
\label{apx:stock_res}

\begin{table*}[ht]
\centering
\resizebox{\textwidth}{!}{%
\begin{tabular}{@{}lccccccc@{}}
\toprule
\textbf{Company (Ticker)} 
& \textbf{Zero-Shot} 
& \textbf{Instruction-Based} 
& \textbf{CoT} 
& \textbf{One-Shot} 
& \multicolumn{2}{c}{\textbf{Self-Verification}} 
& \textbf{Average Observation} \\
\cmidrule(lr){6-7}
 &  &  &  &  & \textbf{Without hint} & \textbf{With hint} & \\
\midrule
\textbf{Google Alphabet (GOOGL)} 
& 64.76\% 
& 23.25\% 
& 55.80\% 
& 39.91\% 
& 17.65\% 
& 27.45\% 
& 212.0 ± 33.0 \\
\textbf{Amazon (AMZN)}           
& 66.81\% 
& 31.72\% 
& 51.11\% 
& 48.68\% 
& 32.95\% 
& 21.39\% 
& 216.4 ± 24.3 \\
\textbf{Apple (AAPL)}            
& 78.88\% 
& 39.84\% 
& 72.91\% 
& 53.39\% 
& 27.13\% 
& 16.60\% 
& 250.2 ± 1.8 \\
\textbf{Meta (META)}             
& 69.48\% 
& 13.02\% 
& 60.85\% 
& 34.11\% 
& 7.28\% 
& 3.97\% 
& 201.0 ± 28.0 \\
\textbf{Microsoft (MSFT)}        
& 69.37\% 
& 7.66\% 
& 56.76\% 
& 35.14\% 
& 17.99\% 
& 21.69\% 
& 215.4 ± 14.8 \\
\textbf{Nvidia (NVDA)}           
& --      
& --      
& --      
& --      
& -- 
& -- 
& -- \\
\textbf{Tesla (TSLA)}            
& 67.92\% 
& 21.16\% 
& 59.17\% 
& 33.33\% 
& 24.55\% 
& 7.59\% 
& 237.0 ± 7.3 \\
\bottomrule
\end{tabular}%
}
\caption{\small{Leakage Rate (\%) Comparison of Prompting Strategies for Ex-Ante Stock Price Prediction Using Claude-3.5-Sonnet across major tech companies.}}
\label{tab:claude-results}
\end{table*}

\begin{table*}[ht]
\centering
\resizebox{\textwidth}{!}{%
\begin{tabular}{@{}lccccccc@{}}
\toprule
\textbf{Company (Ticker)} 
& \textbf{Zero-Shot} 
& \textbf{Instruction-Based} 
& \textbf{CoT} 
& \textbf{One-Shot} 
& \multicolumn{2}{c}{\textbf{Self-Verification}} 
& \textbf{Average Observation} \\
\cmidrule(lr){6-7}
 &  &  &  &  & \textbf{Without hint} & \textbf{With hint} & \\
\midrule
\textbf{Google Alphabet (GOOGL)} 
& 18.67\% 
& 10.60\% 
& 6.20\% 
& 3.36\% 
& 0.00\% 
& 3.17\% 
& 128.4 ± 34.2 \\
\textbf{Amazon (AMZN)}           
& 30.07\% 
& 0.00\% 
& 1.41\% 
& 0.60\% 
& 1.35\% 
& 2.70\% 
& 138.0 ± 37.6 \\
\textbf{Apple (AAPL)}            
& 49.03\% 
& 1.83\% 
& 4.08\% 
& 10.70\% 
& 0.66\% 
& 0.66\% 
& 197.2 ± 27.2 \\
\textbf{Meta (META)}             
& 10.19\% 
& 0.00\% 
& 1.92\% 
& 5.77\% 
& 2.44\% 
& 0.00\% 
& 93.0 ± 29.1 \\
\textbf{Microsoft (MSFT)}        
& 9.43\% 
& 0.67\% 
& 1.25\% 
& 1.89\% 
& 1.10\% 
& 1.10\% 
& 143.8 ± 29.8 \\
\textbf{Nvidia (NVDA)}           
& --      
& --      
& --      
& --      
& --      
& --      
& -- \\
\textbf{Tesla (TSLA)}            
& 11.00\% 
& 0.00\% 
& 6.52\% 
& 0.89\% 
& 1.37\% 
& 0.00\% 
& 99.2 ± 17.9 \\
\bottomrule
\end{tabular}%
}
\caption{\small{Leakage Rate (\%) Comparison of Prompting Strategies for Ex-Ante Stock Price Prediction Using Gemini-1.5-Pro-002 across major tech companies.}}
\label{tab:gemini-results}
\end{table*}

\begin{table*}[ht]
\centering
\resizebox{\textwidth}{!}{%
\begin{tabular}{@{}lccccccc@{}}
\toprule
\textbf{Company (Ticker)} 
& \textbf{Zero-Shot} 
& \textbf{Instruction-Based} 
& \textbf{CoT} 
& \textbf{One-Shot} 
& \multicolumn{2}{c}{\textbf{Self-Verification}} 
& \textbf{Memorization Count (Out of 251)} \\
\cmidrule(lr){6-7}
 &  &  &  &  & \textbf{Without hint} & \textbf{With hint} & \\
\midrule
\textbf{Google Alphabet (GOOGL)} 
& 60.00\% 
& 3.77\% 
& 2.59\% 
& 41.55\% 
& 0.85\% 
& 2.56\% 
& 180.0 ± 40.7 \\
\textbf{Amazon (AMZN)}           
& 62.72\% 
& 3.11\% 
& 0.00\% 
& 39.27\% 
& 2.92\% 
& 4.38\% 
& 208.0 ± 40.0 \\
\textbf{Apple (AAPL)}            
& 84.21\% 
& 2.87\% 
& 1.61\% 
& 47.12\% 
& 0.92\% 
& 4.59\% 
& 217.8 ± 31.2 \\
\textbf{Meta (META)}             
& 59.22\% 
& 2.33\% 
& 7.21\% 
& 40.54\% 
& 0.85\% 
& 6.78\% 
& 196.6 ± 44.5 \\
\textbf{Microsoft (MSFT)}        
& 66.67\% 
& 0.45\% 
& 0.86\% 
& 42.02\% 
& 1.39\% 
& 0.00\% 
& 179.2 ± 51.3 \\
\textbf{Nvidia (NVDA)}           
& --      
& --      
& --      
& --      
& --      
& --      
& -- (no data) \\
\textbf{Tesla (TSLA)}            
& 65.80\% 
& 1.37\% 
& 0.44\% 
& 43.28\% 
& 2.15\% 
& 2.15\% 
& 205.0 ± 17.0 \\
\bottomrule
\end{tabular}%
}
\caption{\small{Leakage Rate (\%) Comparison of Prompting Strategies for Ex-Ante Stock Price Prediction Using GPT-4o-2024-08-06 across major tech companies (values in \%).}}
\label{tab:gpt4-results}
\end{table*}

\begin{table*}[ht]
\centering
\begin{tabular}{lccc}
\toprule
\textbf{Model} & \textbf{Without hint} & \textbf{With hint} & \textbf{Memorization Count (Out of 251)} \\
\midrule
Claude-3-5-Sonnet-20241022 & 21.26\% & 16.45\% &189.50 $\pm$ 38.97\\
Gemini-1.5-Pro-002         &  1.15\% &  1.27\% &82.17 $\pm$ 37.50\\
GPT-4o-2024-08-06          &  1.51\% &  3.41\% &153.33 $\pm$ 40.44\\
\bottomrule
\end{tabular}
\caption{Average leakage rates for Self-Verification (In Conversation)}
\label{tab:gen-validate-in-context}
\end{table*}

\begin{table*}[ht]
\centering
\begin{tabular}{lccc}
\toprule
\textbf{Model} & \textbf{Without hint} & \textbf{With hint} & \textbf{Memorization Count (Out of 251)} \\
\midrule
Claude-3-5-Sonnet-20241022 & 0.11\% & 0.77\% & 189.67 $\pm$ 39.01 \\
Gemini-1.5-Pro-002         & 0.73\% & 1.76\% &  89.25 $\pm$ 32.70 \\
GPT-4o-2024-08-06          & 19.66\% & 4.12\% & 156.50 $\pm$ 38.85 \\
\bottomrule
\end{tabular}
\caption{\small{Average leakage rates for Self-Verification (Independent setting)}}
\label{tab:gen-validate-context-free}
\end{table*}



\begin{figure}[t]
    \centering
    \includegraphics[width=\linewidth]{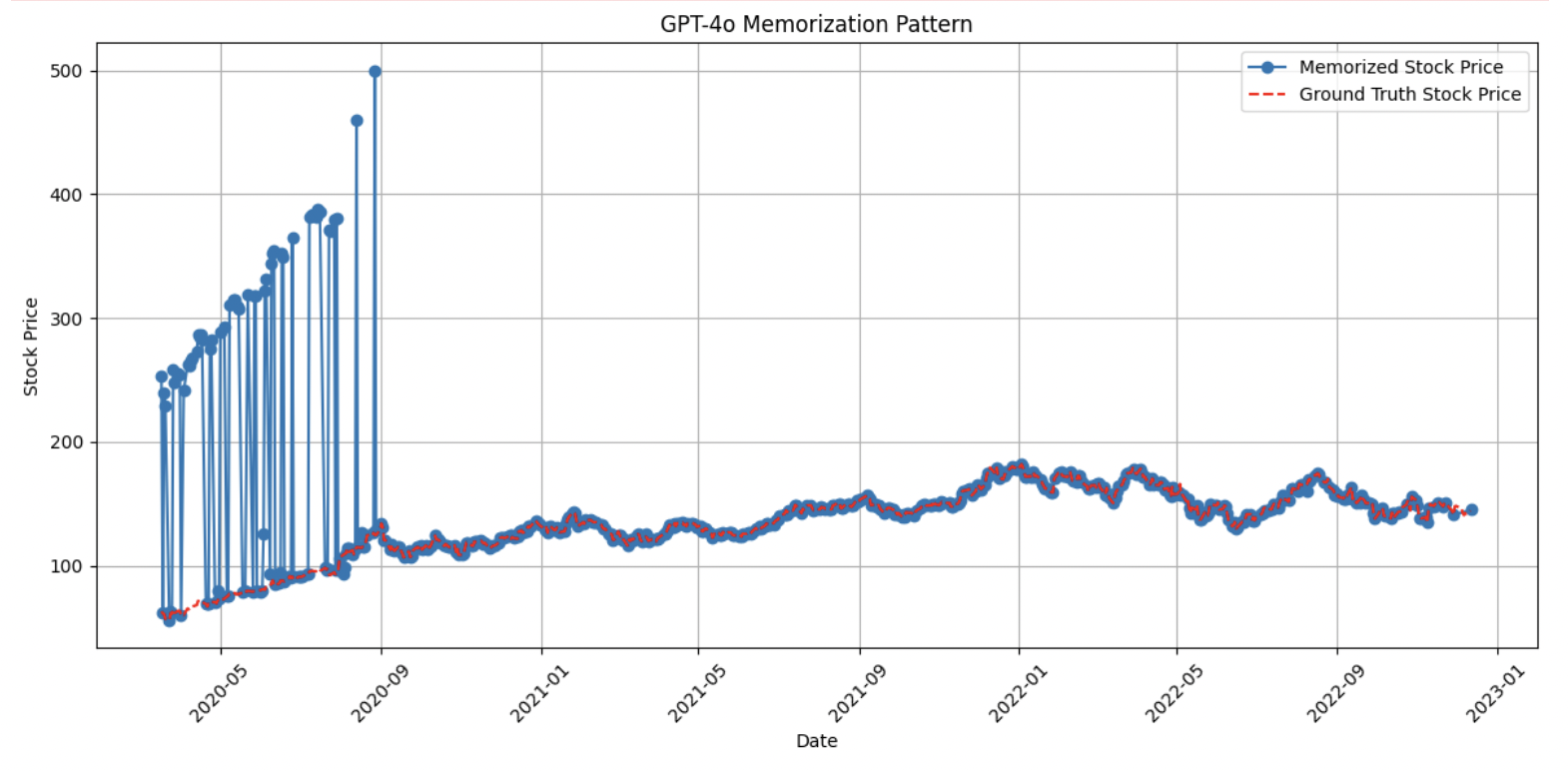}
    \caption{GPT-4o's historical stock price memorization pattern for AAPL. The blue line represents model-predicted prices while the red dashed line shows the ground truth historical prices. The plot demonstrates significantly improved memorization accuracy post-2021, forming a natural temporal boundary for our ExAnte analysis.}
    \label{fig:stock_memory}
\end{figure}

Figure \ref{fig:stock_memory} illustrates the memorization patterns of stock prices for AAPL across different time periods. The model exhibits significant volatility and inconsistent memorization of stock prices prior to mid-2020, as evidenced by the erratic blue line fluctuations. However, post-2021, the model demonstrates markedly improved stability in price memory, with predictions closely tracking the actual stock prices (shown by the dashed line). This empirical observation informed our decision to focus the stock prediction task on the post-2021 period, where the model's memorization behavior shows greater consistency and reliability.

Tables \ref{tab:claude-results}, \ref{tab:gemini-results}, and \ref{tab:gpt4-results} provide company-level breakdowns of leakage rates across different prompting strategies for Claude-3.5-Sonnet, Gemini-1.5-Pro-002, and GPT-4o-2024-08-06, respectively. The data shows that Zero-Shot prompting consistently yields the highest leakage rates across all models (ranging from 64-84\% for both Claude and GPT-4o, and 9-49\% for Gemini). In contrast, the Self-Verification strategy demonstrates the most effective containment of future information across all models, particularly when implemented with hints. These detailed results align with and further substantiate the aggregate findings presented in Table 2.



\subsection{Publication}
For the Publication dataset, we calculated the data leakage rate at the query level, which is the average atomic claim leakage rate per query. The results are shown in table \ref{tab:Pub_dataset_query_level_Leakage_Rate}. It can be observed that all models exhibit leakage rates of about 30-40\%, except for the Self-verification prompt strategy. Self-verification prompting significantly reduces the leakage rates of Claude and Gemini to below 20\%, but not GPT-4o, which remains the same as other prompt strategies. This result is similar to the findings in table \ref{tab:pub_combined}.

\begin{table*}[ht]
\centering
\small
\resizebox{\textwidth}{!}{
\begin{tabular}{lccccc}
\hline
\textbf{Model} 
& \textbf{Zero-shot} 
& \textbf{Instruction-based } 
& \textbf{Chain-of-thought} 
& \textbf{One-shot} 
& \textbf{Self-Verification} \\
\hline
\textbf{GPT-4o   }        & 41.84 & 33.67 & 37.76 & 34.69 & 34.02 \\
\textbf{Claude-3.5-sonnet }   & 32.65 & 33.68 & 37.23 & 32.99 & 10.11 \\
\textbf{Gemini-1.5-pro}       & 35.05 & 34.07 & 35.48 & 34.29 & 18.06 \\
\hline
\end{tabular}
}
\caption{Average query level leakage rates (\%) across different models and prompting strategies in Publication dataset. Here, the 50\% is a natural baseline as discussed in the main paper Section\ref{3.4publication}.}
\label{tab:Pub_dataset_query_level_Leakage_Rate}
\end{table*}

\section{Self-Verification Prompting Analysis}\label{app:self-ver-analysis}
As the most effective prompting strategy, self-verification prompting warrants a more comprehensive analysis. We aim to examine its impact across different datasets, investigate its limitations, and explore why it performs poorly in certain cases, such as on the Wikipedia dataset.




\subsection{Failure Modes of Self-Verification Prompting}

Self-verification prompting aims to enhance temporal adherence by prompting the model to reassess and regenerate its response when necessary. However, as the model is not explicitly informed whether its original response contains leakage, its ability to self-correct varies. Below, we outline the primary failure modes observed across the four datasets.

\subsubsection{Missed Leakage (Failure to Detect Leakage)}
The model generates a response that contains post-cutoff knowledge but fails to recognize this during self-verification. As a result, it confirms its original response without modification, leaving the leakage uncorrected. 
GPT-4o suffers from missed leakage in the Publication dataset: among 294 self-verification trials (98 samples in the dataset with three repeated experiments), there are 89 failed responses as in \ref{app:d1.1.4_failed_response}, 161 "no leakage" and 44 "has leakage" while the actual leakage rate is 80\%.

Example (Publication Dataset):\\
\noindent\rule{0.5\linewidth}{0.4pt}\\
\textit{Leaked Publication:} BERT: Pre-training of Deep Bidirectional Transformers for Language Understanding. \\
\textit{Cutoff Date:} 2018-07-01 \\
\textit{Ground Truth:} The earliest accessible date for BERT is 2018-10-11, which is after the cutoff. \\
\textit{Self-Verification Response:} "No leakage." \\
\noindent\rule{0.5\linewidth}{0.4pt}\\
Possible Cause: The model lacks a clear mechanism to differentiate between pre-cutoff and post-cutoff knowledge, especially when factual recall is strong.

\subsubsection{Ineffective Regeneration (Leakage Persists in Revised Response)}
The model detects potential leakage and attempts to revise its response, but the regenerated output still includes post-cutoff information, often reformulated rather than removed.

Example (Wikipedia Dataset): \\
\noindent\rule{0.5\linewidth}{0.4pt}\\
\textit{Leaked Claim:} "Facebook went public with an initial public offering (IPO) in May 2012." \\
\textit{Cutoff Date:} 2011-12-31 \\
\textit{Ground Truth:} Pre-cutoff content only speculated about a potential IPO by 2013, with no knowledge of the actual IPO date or outcome. \\
\textit{Self-Verification Response:} "Facebook's successful 2012 IPO raised \$16 billion." \\
\noindent\rule{0.5\linewidth}{0.4pt}\\
Possible Cause: The model does not effectively filter post-cutoff knowledge, leading to superficial modifications that fail to correct the issue.

\subsubsection{Overcorrection (False Positive Leading to New Leakage)}
The model wrongly flags its original response as containing leakage when it was actually valid. In revising its answer, it introduces real leakage.

Example (Wikipedia Dataset):  \\
\noindent\rule{0.5\linewidth}{0.4pt}\\
\textit{Original Claim:} "Facebook's acquisition of Instagram marked its expansion into photo-sharing platforms." \\
\textit{Cutoff Date:} 2011-12-31 \\
\textit{Ground Truth:} Pre-cutoff content confirms Facebook’s interest in Instagram, but the acquisition had not yet occurred. \\
\textit{Self-Verification Response:} "Facebook's \$1 billion strategic acquisition in April 2012 successfully expanded its social presence." \\
\noindent\rule{0.5\linewidth}{0.4pt}\\
Possible Cause: The model struggles to differentiate between valid temporal reasoning and accidental memorization, leading it to reject legitimate responses.

\subsubsection{Failed Response (No Regeneration, Original Leakage Persists)}
\label{app:d1.1.4_failed_response}
After self-verification, the model either repeats the same answer or refuses to generate an alternative, leaving the original leakage uncorrected.
In the independent setting, the Gemini-1.5-Pro has a high failure rate of  39.15\%  and 43.61\% (without hint and with hint) while other models low.\\
Example (Stock Dataset):  \\
\noindent\rule{0.5\linewidth}{0.4pt}\\
\textit{Leaked Prediction:} "256.06" (Microsoft stock price on 2022-09-07) \\
\textit{Cutoff Date:} 2021-09-07 \\
\textit{Ground Truth:} The actual stock price on 2022-09-07 was "258.09." \\
\textit{Self-Verification Response:} "I cannot predict future stock prices or provide a hypothetical prediction without using information beyond 2021-09-07. Therefore, I will maintain my previous response." \\
\noindent\rule{0.5\linewidth}{0.4pt}\\
Possible Cause: The model lacks a robust self-correction mechanism, leading to cases where it cannot confidently generate a revised response.

\subsubsection{Additional Observations}
\begin{itemize}
    \item \textit{Self-verification prompting does not guarantee correction.} The model’s ability to detect and fix leakage remains inconsistent, leading to many cases where leakage is left unchanged.
    \item \textit{Failure patterns vary across datasets.} Open-ended tasks (Wikipedia, Publications) exhibit more persistent leakage due to difficulty in verifying event timelines, whereas numerical tasks (Stock) suffer more from overcorrection.
    \item \textit{Regeneration can reinforce mistakes.} In cases where the model falsely detects leakage, its revised responses sometimes introduce new errors instead of fixing existing ones.
\end{itemize}

These findings indicate that while self-verification prompting helps enforce temporal constraints, it is not a complete solution. Future research should explore improved verification mechanisms such as external fact-checking, iterative multi-turn validation, or reinforcement-based feedback.

\subsection{In-Conversation vs. Independent Self-Verification}

The effectiveness of Self-Verification differs significantly between in-conversation (where the model reassesses its own response) and independent verification (where an identical model, without prior context, evaluates the response). Tables \ref{tab:gen-validate-in-context} and \ref{tab:gen-validate-context-free} reveal that models generally exhibit lower leakage rates in the independent verification setting compared to the in-conversation setting.

Claude-3.5-Sonnet demonstrates the largest disparity, with leakage rates dropping from 21.26\% (without hint) and 16.45\% (with hint) in the in-conversation setting to 0.11\% and 0.77\% in the independent setting. This suggests that maintaining prior conversational context may interfere with the model’s ability to filter post-cutoff knowledge effectively. Similarly, Gemini-1.5-Pro maintains notably lower leakage rates in the independent setting (0.73\%-1.76\%) compared to in-conversation (1.15\%-1.27\%), indicating that removing prior context enhances its ability to adhere to temporal constraints.

GPT-4o exhibits the most stable performance across both settings but still shows improvements in the independent setting, particularly when hints are included (leakage drops from 3.41\% in in-conversation to 4.12\% in independent verification). These patterns suggest that removing conversational context helps models better contain future information during Self-Verification. The performance gap highlights the potential influence of implicit contextual priming, where models anchored to prior responses struggle to reassess their outputs independently. This raises important considerations for designing effective self-verification frameworks, where a fully independent judger may yield stricter adherence to temporal constraints than one operating within a multi-turn conversation.

\subsection{The Prompt Content: With Hint vs. Without Hint}

The effectiveness of Self-Verification is influenced by whether the model is provided with an explicit hint regarding ground truth information. Tables \ref{tab:gen-validate-in-context} and \ref{tab:gen-validate-context-free} reveal that incorporating hints reduces information leakage in specific scenarios, though the impact varies across models.

In the in-conversation setting (Table \ref{tab:gen-validate-in-context}), Claude-3.5-Sonnet exhibits a notable reduction in leakage rate from 21.26\% to 16.45\% when hints are provided. However, Gemini-1.5-Pro shows minimal change, suggesting that the hint does not strongly influence its verification process. In contrast, GPT-4o demonstrates a slight increase in leakage (from 1.51\% to 3.41\%), indicating a potential overcorrection effect where exposure to hints may inadvertently reinforce reliance on post-cutoff knowledge.

The independent setting (Table \ref{tab:gen-validate-context-free}) follows a similar trend. GPT-4o experiences a dramatic reduction in leakage when hints are included (from 19.66\% to 4.12\%), suggesting that explicit guidance significantly enhances its ability to self-regulate. Meanwhile, Claude-3.5-Sonnet exhibits a more modest improvement (from 0.11\% to 0.77\%), and Gemini-1.5-Pro shows a slight increase in leakage, indicating that hints may introduce unintended biases rather than always reinforcing adherence to pre-cutoff knowledge.

These findings suggest that while hints can improve Self-Verification performance by reinforcing temporal constraints, their effectiveness depends on the model and context. In some cases, hints lead to beneficial correction, whereas in others, they introduce overcorrection or fail to provide meaningful improvements. Understanding how different models process hints is essential for designing robust self-verification frameworks.



\section{Other}
\subsection{Model Versions and Inference Configurations}
\label{app:models_versions}

We evaluate the following model versions in our benchmark:
\begin{itemize}
    \item \textbf{GPT-4o} (\texttt{gpt-4o-2024-08-06}): Released August 6, 2024
    \item \textbf{Claude 3.5 Sonnet} (\texttt{claude-3-5-sonnet-20241022}): Released October 22, 2024
    \item \textbf{Gemini 1.5 Pro} (\texttt{gemini-1.5-pro-002}): September 24, 2024
\end{itemize}

We apply consistent decoding parameters per task:
\begin{itemize}
    \item \textbf{Wikipedia}: temperature = 0.7, top-p = default, max tokens = 500
    \item \textbf{Stock}: temperature = 0.0, top-p = default, max tokens = 1000
    \item \textbf{QA and Publication}: temperature = 1.0, top-p = default, max tokens = 1000
\end{itemize}

\section{Limitations}

While our study systematically evaluates temporal leakage, it does not assess the factual correctness of model responses. This raises the possibility that models could minimize leakage by generating unverifiable or hallucinated outputs instead of adhering to pre-cutoff constraints. Although our results suggest that models do not rely on such strategies—given the observed leakage rates—developing complementary metrics that jointly measure factual correctness and temporal consistency remains an important direction for future work.  

Additionally, our benchmark focuses on evaluating leakage across a limited number of models and prompting strategies. Future studies could extend this analysis to fine-tuned models, retrieval-augmented approaches, or architectural modifications explicitly designed to enhance temporal adherence.  

\section{Impact}
Our enforces strict temporal cutoffs to prevent large language models from “peeking” at future data. It can potentially improve trust in time‑sensitive domains like finance or historical research. One potential negative impact is that one could use the data to fine‑tune models that conceal their use of future knowledge. We suggest using our dataset mainly for testing purposes.


\newpage


\newpage
\section*{NeurIPS Paper Checklist}

The checklist is designed to encourage best practices for responsible machine learning research, addressing issues of reproducibility, transparency, research ethics, and societal impact. Do not remove the checklist: {\bf The papers not including the checklist will be desk rejected.} The checklist should follow the references and follow the (optional) supplemental material.  The checklist does NOT count towards the page
limit. 

Please read the checklist guidelines carefully for information on how to answer these questions. For each question in the checklist:
\begin{itemize}
    \item You should answer \answerYes{}, \answerNo{}, or \answerNA{}.
    \item \answerNA{} means either that the question is Not Applicable for that particular paper or the relevant information is Not Available.
    \item Please provide a short (1–2 sentence) justification right after your answer (even for NA). 
\end{itemize}

{\bf The checklist answers are an integral part of your paper submission.} They are visible to the reviewers, area chairs, senior area chairs, and ethics reviewers. You will be asked to also include it (after eventual revisions) with the final version of your paper, and its final version will be published with the paper.

The reviewers of your paper will be asked to use the checklist as one of the factors in their evaluation. While "\answerYes{}" is generally preferable to "\answerNo{}", it is perfectly acceptable to answer "\answerNo{}" provided a proper justification is given (e.g., "error bars are not reported because it would be too computationally expensive" or "we were unable to find the license for the dataset we used"). In general, answering "\answerNo{}" or "\answerNA{}" is not grounds for rejection. While the questions are phrased in a binary way, we acknowledge that the true answer is often more nuanced, so please just use your best judgment and write a justification to elaborate. All supporting evidence can appear either in the main paper or the supplemental material, provided in appendix. If you answer \answerYes{} to a question, in the justification please point to the section(s) where related material for the question can be found.

IMPORTANT, please:
\begin{itemize}
    \item {\bf Delete this instruction block, but keep the section heading ``NeurIPS Paper Checklist"},
    \item  {\bf Keep the checklist subsection headings, questions/answers and guidelines below.}
    \item {\bf Do not modify the questions and only use the provided macros for your answers}.
\end{itemize}


\begin{enumerate}

\item {\bf Claims}
    \item[] Question: Do the main claims made in the abstract and introduction accurately reflect the paper's contributions and scope?
    \item[] Answer: \answerYes{} 
    \item[] Justification: The main claims made in the abstract and introduction accurately reflect the paper's contributions and scope

\item {\bf Limitations}
    \item[] Question: Does the paper discuss the limitations of the work performed by the authors?
    \item[] Answer: \answerYes{} 
    \item[] Justification: We provided a limitation section in Appendix.

\item {\bf Theory assumptions and proofs}
    \item[] Question: For each theoretical result, does the paper provide the full set of assumptions and a complete (and correct) proof?
    \item[] Answer: \answerYes{} 
    \item[] Justification: Although our paper is not theoretical work, we provide the full set of assumptions of our problems.

    \item {\bf Experimental result reproducibility}
    \item[] Question: Does the paper fully disclose all the information needed to reproduce the main experimental results of the paper to the extent that it affects the main claims and/or conclusions of the paper (regardless of whether the code and data are provided or not)?
    \item[] Answer: \answerYes{} 
    \item[] Justification: We disclose all the infomation needed to reproduce the main experimental results.

\item {\bf Open access to data and code}
    \item[] Question: Does the paper provide open access to the data and code, with sufficient instructions to faithfully reproduce the main experimental results, as described in supplemental material?
    \item[] Answer: \answerYes{} 
    \item[] Justification: We provide the open access to the data and code.

\item {\bf Experimental setting/details}
    \item[] Question: Does the paper specify all the training and test details (e.g., data splits, hyperparameters, how they were chosen, type of optimizer, etc.) necessary to understand the results?
    \item[] Answer: \answerYes{} 
    \item[] Justification: We specify all the training and test details necessary to understand the results

\item {\bf Experiment statistical significance}
    \item[] Question: Does the paper report error bars suitably and correctly defined or other appropriate information about the statistical significance of the experiments?
    \item[] Answer: \answerNo{} 
    \item[] Justification: As the experiments rely on the API calls to multiple LLMs, it is very expensive to repeat the experiments multiple times. Instead of reporting error bars, we tested each method on multiple datasets.

\item {\bf Experiments compute resources}
    \item[] Question: For each experiment, does the paper provide sufficient information on the computer resources (type of compute workers, memory, time of execution) needed to reproduce the experiments?
    \item[] Answer: \answerYes{} 
    \item[] Justification: We provide the computer resources needed.
    
\item {\bf Code of ethics}
    \item[] Question: Does the research conducted in the paper conform, in every respect, with the NeurIPS Code of Ethics \url{https://neurips.cc/public/EthicsGuidelines}?
    \item[] Answer: \answerYes{} 
    \item[] Justification: We follow the NeurIPs Code of Ethics.

\item {\bf Broader impacts}
    \item[] Question: Does the paper discuss both potential positive societal impacts and negative societal impacts of the work performed?
    \item[] Answer: \answerYes{} 
    \item[] Justification: We have a section dicuss the potential impaces of our work.
    
\item {\bf Safeguards}
    \item[] Question: Does the paper describe safeguards that have been put in place for responsible release of data or models that have a high risk for misuse (e.g., pretrained language models, image generators, or scraped datasets)?
    \item[] Answer: \answerYes{} 
    \item[] Justification: We describe this in the impact section.

\item {\bf Licenses for existing assets}
    \item[] Question: Are the creators or original owners of assets (e.g., code, data, models), used in the paper, properly credited and are the license and terms of use explicitly mentioned and properly respected?
    \item[] Answer: \answerYes{} 
    \item[] Justification: The creators or original owners of assets used in the paper, are properly credited and the license and terms of use are explicitly mentioned and properly respected.

\item {\bf New assets}
    \item[] Question: Are new assets introduced in the paper well documented and is the documentation provided alongside the assets?
    \item[] Answer: \answerYes{} 
    \item[] Justification: The new assets introduced in the paper are documented.

\item {\bf Crowdsourcing and research with human subjects}
    \item[] Question: For crowdsourcing experiments and research with human subjects, does the paper include the full text of instructions given to participants and screenshots, if applicable, as well as details about compensation (if any)? 
    \item[] Answer: \answerNA{} 
    \item[] Justification: The paper does not involve crowdsourcing nor research with human subjects.

\item {\bf Institutional review board (IRB) approvals or equivalent for research with human subjects}
    \item[] Question: Does the paper describe potential risks incurred by study participants, whether such risks were disclosed to the subjects, and whether Institutional Review Board (IRB) approvals (or an equivalent approval/review based on the requirements of your country or institution) were obtained?
    \item[] Answer: \answerNA{} 
    \item[] Justification: The paper does not involve crowdsourcing nor research with human subjects.

\item {\bf Declaration of LLM usage}
    \item[] Question: Does the paper describe the usage of LLMs if it is an important, original, or non-standard component of the core methods in this research? Note that if the LLM is used only for writing, editing, or formatting purposes and does not impact the core methodology, scientific rigorousness, or originality of the research, declaration is not required.
    \item[] Answer: \answerYes{} 
    \item[] Justification: We describe the usage of LLM.

\end{enumerate}

\end{document}